\documentclass{article}
\usepackage{spconf,amsmath,graphicx}
\usepackage{amsmath,bm}
\usepackage[ruled,vlined,linesnumbered]{algorithm2e}
\usepackage{multirow}
\usepackage{pgfplots}
\pgfplotsset{compat=1.17}
\usepackage[colorlinks=true,citecolor=magenta,linkcolor=magenta]{hyperref}
\usepackage{comment}
\usepackage{enumitem}

\title{IKD+: Reliable Low Complexity Deep Models for Retinopathy Classification}
\name{%
\begin{tabular}{@{}c@{}}
Shreyas Bhat Brahmavar$^{1}$,
Rohit Rajesh$^{1}$,
Tirtharaj Dash$^{2}$,
Lovekesh Vig$^{3}$,
Tanmay Tulsidas Verlekar$^{1}$,\\
Md Mahmudul Hasan$^{4}$,
Tariq Khan$^{4}$,
Erik Meijering$^{4}$,
Ashwin Srinivasan$^{1}$
\end{tabular}}
\address{$^{1}$ APPCAIR, BITS Pilani, K.K. Birla Goa Campus, India\\
        $^{2}$ Boolean Lab, University of California, San Diego, USA\\
        $^{3}$ TCS Research, India\\
        $^{4}$ School of CSE, UNSW, Sydney, Australia}
   
\begin{document}
\maketitle
\begin{abstract}
Deep neural network (DNN) models for retinopathy have estimated predictive accuracies in the mid-to-high $90\%$. However, the following aspects remain unaddressed: State-of-the-art models are complex and require substantial computational infrastructure to train and deploy; The reliability of predictions can vary widely.
In this paper, we focus on these aspects and propose a form of iterative knowledge distillation (IKD), called IKD+ that incorporates a tradeoff between size, accuracy and reliability. 
We investigate the functioning of IKD+ using two widely used techniques for estimating model calibration (Platt-scaling and temperature-scaling), using the best-performing model available, which is an ensemble of EfficientNets with approximately 100M parameters. 
We demonstrate that IKD+ equipped with temperature-scaling results in models that show up to approximately 500-fold decreases in the number of parameters than the original ensemble without a significant loss in accuracy. In addition, calibration scores (reliability) for the IKD+ models are as good as or better than the base model.
\end{abstract}

\section{Introduction}
\label{sec:intro}

Retinopathy is a disease that affects the retina, the light-sensitive layer of tissue at the back of the eye, with number of cases
predicted to be around 191 million by 2030~\cite{zheng2012worldwide}. Usually,
it is a common complication of diabetes and can also be caused by other conditions, such as glaucoma and age-related macular degeneration. If left undiagnosed or untreated, retinopathy can lead to vision loss, and even blindness \cite{wong_diabetic_2016,10.1001/jama.298.8.944,mohamed2007management}.
With remarkable progress in artificial intelligence (AI)---especially, Deep Neural Networks (DNNs)---
its successful application in healthcare there is enormous potential
to adopt it in diagnosing retinopathy. Furthermore,
deep learning models for retinopathy have achieved high predictive accuracies, with some models achieving performance in the mid-to-high 90\% range~\cite{Gargeya2017,9740851,muller2021multi}. However, these models are often highly complex, requiring substantial computational infrastructure and expertise to train and deploy in real-world usecases~\cite{muller2021multi}. Additionally, the reliability of their predictions can vary widely, making it difficult to use them confidently in clinical practice. 
Furthermore, these models are often black boxes, providing little or no meaningful explanations for their predictions, which can be especially problematic when the complexity of the model is high.

In this paper, we address these challenges by exploring iterative knowledge distillation \cite{yalburgi2020empirical} that results in progressively simpler deep-learning models for retinopathy. We base our work on advanced model calibration \cite{guo2017calibration} techniques, which  show that it is possible to improve the reliability of high-probability predictions without sacrificing predictive accuracy. We demonstrate our approach using two state-of-the-art methods for multi-disease retinopathy classification. Our results show that significantly simpler models can achieve similar predictive accuracies to the original models. Furthermore, the calibration scores of the simpler models are as good as or better than those of the more complex models.

Our approach has the potential to significantly improve the diagnosis (and treatment) of retinopathy by providing more reliable, low-complexity deep learning models that can be easily trained and deployed in clinical settings. It could also improve the transparency and interpretability of these models, enabling medical professionals to understand the results better and trust their predictions. By addressing these challenges, our approach has the potential to significantly improve the quality of care for millions of individuals throughout the world who suffer from preventable or undetected blindness and vision impairment.
Apart from contributing to the diagnosis and treatment of retinopathy, we also extend the calibration technique to estimate the reliability of models for multi-label classification. The state-of-the-art calibration techniques primarily focus on binary- or multi-class classification problems. Relatively less has been done on multi-label classification problems, as reported in the following sections.

\subsection{Related Works}
Lyu et al.~\cite{9871762} proposed a multi-expert deep learning network for classifying retinal fundus images based on disease. The network comprises multiple classifiers whose logits are stacked to take an ensemble for the final prediction. 
In Müller et al. \cite{muller2021multi}, the authors propose a multi-classifier approach for classifying retinal images into normal and abnormal classes and identifying specific disease labels among 28 classes. The approach utilizes ensemble learning to combine the predictive capabilities of several heterogeneous deep convolutional neural network models. The authors demonstrate their pipeline's high accuracy and comparability with other state-of-the-art pipelines for retinal disease prediction. There are comprehensive surveys 
on application of deep learning for retinopathy diagnosis~\cite{asiri2019deep,atwany2022deep}.

Many of the deep learning based techniques use complex 
deep network structures that often require heavy computational
resource for training and deployment, for example,
models built with DenseNet201, EfficientNetB4. Some
techniques also use ensembles of ML models, such as stacked logistic regression, which can further increase the computational requirements. A further limitation of the state-of-the-art method is in the reliability of their predictions. Most models use a stacking setup, where a binary logistic regression algorithm is applied for each class individually, resulting in 29 distinct logistic regression models (for the problem being
solved in this paper). However, how well these models are calibrated and whether they produce reliable predictions is unclear. Calibration refers to the degree to which predicted probabilities match the true outcomes. In the medical field, it is essential to ensure that the models make predictions with a higher degree of reliability. Therefore, we evaluate the calibration of the proposed models using Expected Calibration Error \cite{guo2017calibration,patra2023calibrating}.

\subsection{Major contributions}
%Deep learning models for retinopathy diagnosis have relied heavily on computationally expensive state-of-the-art methods. However, these methods are unreliable in terms of their predicted probabilities aligning with the true frequencies of the target variable, making them impractical for deployment in real-world settings. 
Our contributions toward the construction of a reliable deep-learning model for retinopathy diagnosis are as follows: 
\begin{itemize}[noitemsep, nolistsep]
    \item We propose a novel method called IKD+ for the accurate transfer of knowledge from a large, complex model (i.e., the teacher) to a smaller, more efficient model (i.e., the student) in an iterative manner;
    \item We employ Platt- and Temperature-scaling to enhance the reliability of the predictions made by the student model;
    \item Through extensive experimentation, we demonstrate the effectiveness of our proposed method in improving the accuracy and reliability of retinopathy diagnosis.
\end{itemize}

The rest of the paper is organised as follows: 
We elaborate on our proposal in Section~\ref{sec:rkd}, present evaluation details in Section~\ref{sec:emp} and discuss our results in ~\ref{sec:res}. %We conclude our findings in Section~\ref{sec:concl}.

% \section{Formatting your paper}
% \label{sec:format}

\section{Reliable IKD (IKD+)}
\label{sec:rkd}

In this section, we introduce the concept of reliable iterative knowledge distillation (IKD+), which aims to identify a simpler variant of a (large, complex) model that (a) has an accuracy comparable to the more complex model, and (b) makes reliable predictions (roughly: one with low prediction error for high-confidence predictions). This latter requirement arises because it is now recognized that modern deep network models are known occasionally to make very poor predictions with very high confidence \cite{guo2017calibration} \cite{Jiang2012}. This can be especially problematic if such predictions are used to provide medical assistance. To achieve these two goals,  we use the method of IKD (described in \cite{yalburgi2020empirical}), with a loss function that accounts for model calibration at each step of the iterative procedure. For completeness, the IKD procedure is shown below: the specifics of the loss function and calibration error follow.
IKD considers a model $M$ as a pair of ($\sigma$, $\bm{\pi}$), where $\sigma$ denotes
the model's structure and refers to the organisation of neurons, layers, and activation, and $\bm{\pi}$ denotes the model's parameters (synaptic weights). Knowledge distillation can be considered a function $KD$ that finds $\bm{\pi}_{j}$ s.t. 
\begin{equation}
    KD((\sigma_{i}, \bm{\pi}_{i}), \cdot) = (\sigma_{j}, \bm{\pi}_{j}). 
\end{equation}

Similar to \cite{yalburgi2020empirical}
we assume a refinement operator $\rho$ that
is applied to a structure of the deep network, 
which results in a new and typically simpler structure.
A simpler model is less wide and/or depth and has fewer parameters.
In our work, which is inspired by Yalburgi \textit{et al.}~\cite{yalburgi2020empirical} 
we propose a refinement ($\rho$) operator that results in \textit{block reduction}, i.e., The number of features within each block in the structure is reduced by some percentage $p$, and; The number of convolution filters in each block is reduced by $p$. The value of $p$ is also set to $0.5$ in our work, following the study in~\cite{yalburgi2020empirical}.

The main contribution of our work is to incorporate the notion of reliability
of prediction by a model into the IKD procedure. For this, we alter the  loss-function
employed in \cite{yalburgi2020empirical} to include the
Expected Calibration Error (ECE) introduced in \cite{guo2017calibration}. We describe
these details in the next subsection.

\subsection*{Calibration-Sensitive Loss:}
We consider two different loss functions called Temperature Scaling and Platt-scaling. In either case, the loss is expressed as: 
\begin{equation}
L\left(M_s \mid D\right)=\frac{1}{m} \sum_{j=1}^m E_j,
\end{equation}
where $E_j$ for Temperature Scaling, $M_s$ is the student model, $D$ is the training data and Platt-scaling can be expressed as (3) and (4), respectively. 
% \begin{figure}[h]

%             \centering
%             \includegraphics[width=0.4\columnwidth]
            
%            $$ L\left(M_s \mid D\right)=\frac{1}{m} \sum_{j=1}^m\left\{2 T^2 \alpha D_{K L}\left(P^{(j)}/T, Q^{(j)}/T\right)-(1-\alpha) \sum_{i=1}^c y_i^{(j)} \log \left(1-\hat{y}_i^{(j)}\right)\right\}$$
% \end{figure}
\begin{equation}
{\footnotesize
2T^2\alpha D_{KL}\left(\frac{P^{(j)}}{T},\frac{Q^{(j)}}{T}\right) + (1-\alpha)\cdot\mathrm{CE(Y,\hat{Y})}
}
\end{equation}
\begin{equation}
{\footnotesize
\alpha D_{K L}\left(WP^{(j)} ,WQ^{(j)  }\right)+(1-\alpha)\cdot\mathrm{CE(Y,\hat{Y})}
}
\end{equation}
In equations (3) and (4), $CE$ refers to Cross-Entropy Loss $D_{KL}$ denotes the KL-divergence \cite{kullback1951information}, and $T$, $W$ are all optimized on the validation set, m denotes the batch size, and c denotes the number of classes. $\alpha$ is the relative importance of the teacher’s guidance w.r.t. hard targets from data. $P
^{(j)}$ and $Q^{(j)}$ are the class probability
distributions from the student and teacher model, respectively, with j
the
data instance; $Y = [y_{1},\ldots,y_{c}]$ is the one-hot encoded ground truth, and
$\hat{Y} = [\hat{y}_{1},\ldots,\hat{y}_{c}]$ is the prediction from the student model.

\begin{algorithm}[htb]
	\SetAlgoLined
	\KwIn{Pre-trained teacher model $M_{0}$; number of iterations $k$; $\sigma_{1}$ for the first student model; Refinement operator $\rho$;}
             
	\KwOut{A compressed model}
	Initialize $i := 1$\;
	\While{$i \leq k$}{
		$M_{i} = \mathsf{IKD+} (M_{i-1}, \sigma_{i}, \cdot)$\; where IKD+ uses a Calibration-Sensitive loss function 
		\label{step:dafl}\;
		$\sigma_{i+1} = \rho(\sigma_{i})$\;
		$increment\ i$\;
	}
	\textbf{return} $M_{k}$\;
	\caption{Proposed IKD+ Procedure}
	\label{proc:ikd+}
\end{algorithm}

\section{Empirical Evaluation}
\label{sec:emp}

Our evaluation aims to investigate the trade-off between predictive accuracy,
reliability, and compression using IKD+. Specifically, we investigate the
following hypotheses for predicting diabetic retinopathy:

\begin{itemize}[noitemsep, nolistsep]
    \item The IKD+ procedure equipped with a suitable estimator of Expected Calibration Error (ECE) constructs more reliable models
        than IKD;
      %  \\, without a significant decrease in predictive performance    over a state-of-the-art prediction; and with model-sizes comparable or better than IKD.%
         \item The use of IKD+ doesn't significantly decrease the predictive performance 
        over state-of-the-art models of comparable size  or models obtained through IKD.
\end{itemize}

\subsection{Dataset and Machines specifications}
\begin{figure}
  \centering
  \includegraphics[width=0.99\linewidth]{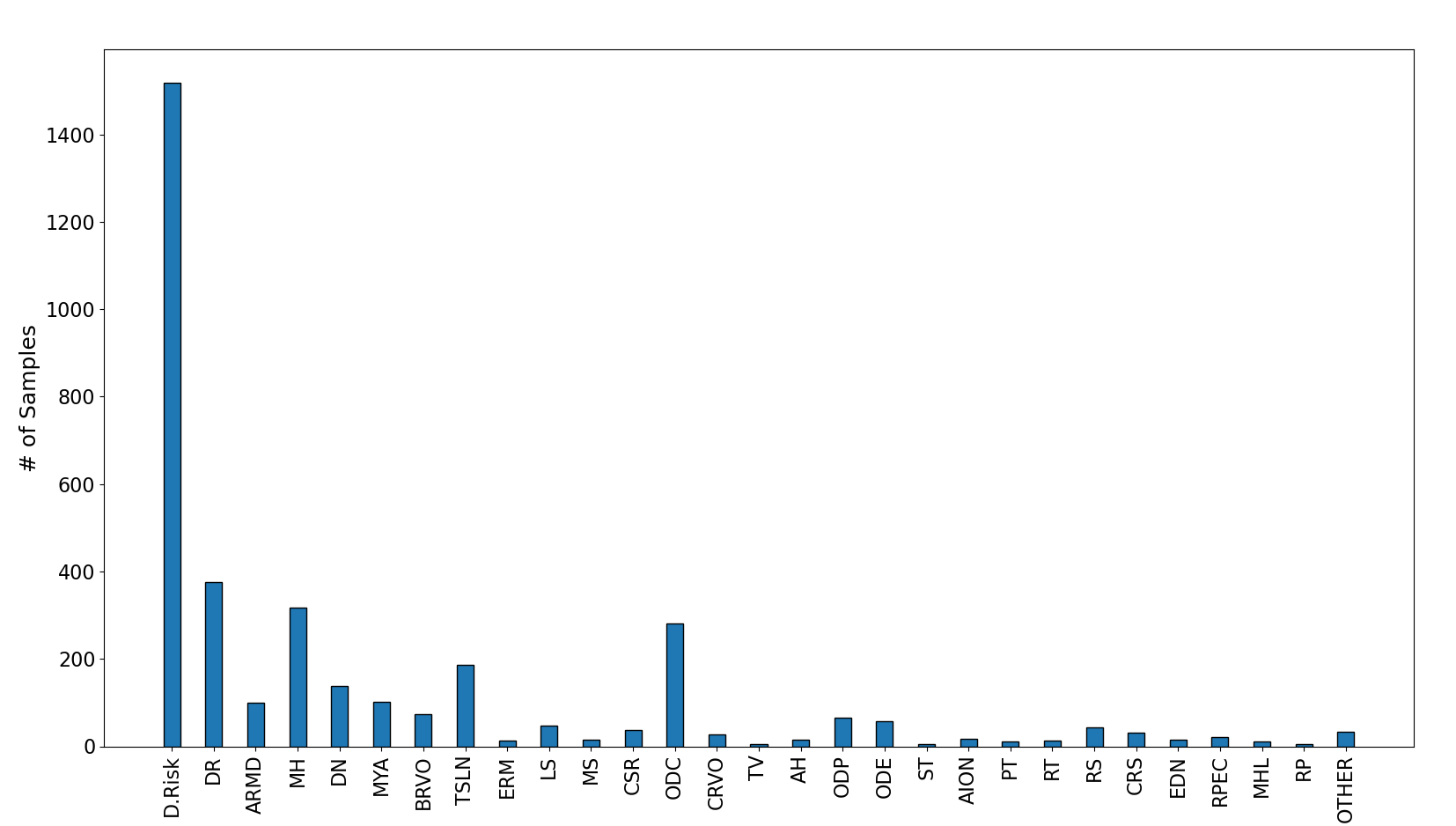}
  \caption{Distribution of the Dataset.}
  \label{fig:dataset}
\end{figure}

We conduct our evaluations on the Retinal Fundus Multi-Disease Image
Dataset \cite{data6020014}. It consists of $3200$ retinal images from which
$1920$ are used for training. A challenge associated with the dataset is the distribution of diseases. While  diabetic retinopathy (DR) contain $376$ examples ($\approx 10\%$), shunt (ST) contains only $5$ ($\approx 0.1\%$). The complete distribution of the dataset is provided in Fig~\ref{fig:dataset}.  All images are in RGB colour format images and $768 \times $768 pixels.

Our proposal, Procedure~\ref{proc:ikd+}, is an improvement of the algorithm presented in \cite{yalburgi2020empirical}. It is implemented in Python on a system containing NVIDIA Tesla V100-SXM2 GPU
with 32GB memory. The code is developed on PyTorch Deep Learning framework \cite{NEURIPS2019_9015}. While the
 expected calibration error (ECE) is calculated using the {\tt{net:cal}} library \cite{Kueppers_2020_CVPR_Workshops}. 

\subsection{Method}
Given a dataset:\\
Split the data into a training set $D_{train}$ and a test set $D_{test}$;\\
        Using $D_{train}$ as training data, % let    $M$ be obtained using a state-of-the-art technique,   with accuracy $A$ and ECE $E$ and size $S$%\\
          set $M$ as the  teacher model with accuracy $A$, ECE $E$ and size $S$;\\
        Let $M_0$ be a baseline model for IKD obtained from $M$, with
            accuracy $A_0$, ECE $E_0$ and size $S_0$ \label{step:m0};
       
        Now for $k = 1,2,\ldots,n$

            \begin{enumerate}[noitemsep, nolistsep]
                \item Let $M_k$ be the model constructed by
                    the IKD procedure presented in \cite{yalburgi2020empirical}
                    and $M_k^+$ be the model constructed by the proposed IKD+ procedure (refer Procedure~\ref{proc:ikd+}). 
             \item Assign the accuracy, ECE and size of $M_k$ be
                $A_k$, $E_k$ and $S_k$ respectively
            \item Similarly, set the accuracy, ECE, and size of $M_k^+$ to
                $A_k^+$, $E_k^+$ and $S_k^+$ respectively
            \item Finally compare: $(A_k,E_k,S_k)$ against $(A,E,S)$, $(A^+,E_k^+,S_k^+)$ against $(A,E,S)$ and $(A_k^+,E_k^+,S_k^+)$ against $(A_k,E_k,S_k)$.
                 
        \end{enumerate}

\noindent
For our evaluation, the following details are relevant:\\
       -- We consider the teacher model to be the winner of the RIADD Grand Challenge \cite{riadd}.
            However, it is an ensemble of EfficientNets~\cite{tan2019efficientnet}. The teacher model $M_0$ for the Step \ref{step:m0} is assigned to the model with the best accuracy in the ensemble, that is, EfficientNet-B6;\\
      -- The evaluation is conducted by setting $k$ to $5$;\\
       Since the dataset is unbalanced, we perform data augmentation to balance the class distribution. The augmentations include rotation, flipping, and changes in brightness, hue and contrast;\\
    -- The $\alpha$ value for the IKD+ procedure is set to $0.7$;\\
     -- The IKD+ is performed with two different techniques: temperature-scaling and Platt-scaling;\\
     -- The IKD and IKD+ employ the same refinement operator $S_k = S_k^+$ for each
        value of $k$;\\
      -- In all cases, we report size in proportion to $M_0$;\\
    -- The models are trained with an Adam optimizer \cite{kingma2015adam}, with a momentum factor $\beta$ of (0.9, 0.999);\\
        -- The hyper-parameters are empirically determined with the batch size set to 16 and the learning rate set to $10^{-4}$.
\section{Results}
\label{sec:res}
%       (A_k+,E_k+,S_k+)  (A_k,E_k,S_k)  (A,[E],S)
% k = 0                                   0.94,[],S
% k = 1
% ....

The results from our experiments are reported in
Tables 1 and 2.  Among the two calibration-sensitive Losses considered, the Temperature scaling performs better than Platt scaling. However, in either case, the accuracy of IKD+ models is better than the IKD model for most values of $k > 0$. The ECE values for IKD+ are also always lower than IKD models. This suggests that the IKD+ models are more reliable than their IKD counterparts, with more confident predictions -see Fig 2.  \\
It should also be noted that compared with the base model $M_0$, the IKD+ results in models with the same accuracy but eight times smaller size. The result is consistent with another experiment where we considered EfficientNet-B5, the second-best model in the ensemble, as our base Model $M_0$ - see Table \ref{tab:efficientnetb5_1}. Thus, it can be concluded that our proposal can achieve accuracy equal to the state-of-the-art with eight $\times$ compression. Even after {$128 \times$} compression, the drop in accuracy is just {$5\%$} - see Table \ref{tab:efficientnetb6_1}. This result is significant when compared to the state-of-the-art ensemble models such as \cite{muller2021multi}, which contain 20 deep convolutional networks. The $M_5$ student model in Table \ref{tab:efficientnetb6_1} achieves compression of {$500 \times$} against such models, while still maintaining an accuracy of $90\%$.
%and within about \textcolor{red}{X\% of}     the state-of-the-art model for values of \textcolor{red}{$k < ...$} (b) The ECE values of IKD+ models are always lower than IKD models for any value     of $k > 0$ (c) The sizes of IKD+ models are the same as IKD models for any value of $k$    and about 500x smaller than the state-of-the-art without significant loss in accuracy%
Tables 3 and 4 report additional results of performing IKD+ with EfficientNet-B5, the second-best model in the ensemble, as the base model. The drop in accuracy is slightly higher than IKD+ with EfficientNet-B6. This drop is expected as the EfficientNet-B5 is a $29 \times$ compressed version of EfficientNet-B6. The results can be used to conclude that IKD+ is better for obtaining small-size models than shifting from EfficientNet-B6 to EfficientNet-B5, with a better trade-off between accuracy, reliability and size.
%, the   The IKD+ results, in this case, are consistent with EfficientNet-B6 results reported in Tables II and III. This suggests that the IKD+ can be successfully used for all the models in the ensemble. 
%(a) For architectures EfficientNet-B6 and EfficientNet-b5, an increase in the iterations results in a decrease in model size, achieving 128-fold compression, respectively. 
%(b) Successive iterations also result in more confident predictions, as shown by the decrease in the ECE values 
\begin{table}[!htb]   
% The baseline comparison model is the initial $M_{0}$. IKD with An entry of S = $M_{j}$ and T = $M_{i}$ for IKD denotes
% that the procedure returned the student model $M_{j}$ after selecting $M_{i}$ as the
% teacher from $M_{0}$, . . . , $M_{i-1}$.
% \begin{minipage}{.5\linewidth}
\centering
\label{tab:efficientnetb61}
\caption{Comparision of IKD+ and IKD using Platt Scaling calibration on EfficientNet-B6}
\resizebox{\columnwidth}{!}{%
\begin{tabular}{cccc|cc|cc|}
\cline{5-8}
                           &                         &                         &      & \multicolumn{2}{c|}{IKD}           & \multicolumn{2}{c|}{IKD+}          \\ \cline{2-8} 
\multicolumn{1}{c|}{} &
  \multicolumn{1}{c|}{Student} &
  \multicolumn{1}{c|}{Teacher} &
  Compression &
  \multicolumn{1}{c|}{ECE} &
  Acc &
  \multicolumn{1}{c|}{ECE} &
  Acc \\ \hline
\multicolumn{1}{|c|}{Base} & \multicolumn{1}{c|}{M0} & \multicolumn{1}{c|}{-}  & 1x   & \multicolumn{1}{c|}{1.95}  & 0.912 & \multicolumn{1}{c|}{0.483} & 0.942 \\ \hline
\multicolumn{1}{|c|}{\multirow{5}{*}{Platt Scaling}} &
  \multicolumn{1}{c|}{M1} &
  \multicolumn{1}{c|}{M0} &
  8x &
  \multicolumn{1}{c|}{1.77} &
  0.917 &
  \multicolumn{1}{c|}{0.472} &
  0.896 \\ \cline{2-8} 
\multicolumn{1}{|c|}{}     & \multicolumn{1}{c|}{M2} & \multicolumn{1}{c|}{M1} & 16x  & \multicolumn{1}{c|}{1.899} & 0.892 & \multicolumn{1}{c|}{0.459} & 0.931 \\ \cline{2-8} 
\multicolumn{1}{|c|}{}     & \multicolumn{1}{c|}{M3} & \multicolumn{1}{c|}{M2} & 32x  & \multicolumn{1}{c|}{1.43}  & 0.88  & \multicolumn{1}{c|}{0.428} & 0.931 \\ \cline{2-8} 
\multicolumn{1}{|c|}{}     & \multicolumn{1}{c|}{M4} & \multicolumn{1}{c|}{M3} & 64x  & \multicolumn{1}{c|}{1.61}  & 0.801 & \multicolumn{1}{c|}{0.412} & 0.889 \\ \cline{2-8} 
\multicolumn{1}{|c|}{}     & \multicolumn{1}{c|}{M5} & \multicolumn{1}{c|}{M4} & 128x & \multicolumn{1}{c|}{1.823} & 0.797 & \multicolumn{1}{c|}{0.412} & 0.844 \\ \hline
\end{tabular}
}
% \end{minipage} 
\end{table}
% \usepackage{multirow}

\begin{table}[!htb]
\centering
\caption{Comparision of IKD+ and IKD using Temperature Scaling calibration on EfficientNet-B6}
\label{tab:efficientnetb6_1}
\resizebox{\columnwidth}{!}{%
\begin{tabular}{cccc|cc|cc|}
\cline{5-8}
                           &                         &                         &      & \multicolumn{2}{c|}{IKD}           & \multicolumn{2}{c|}{IKD+}          \\ \cline{2-8} 
\multicolumn{1}{c|}{} &
  \multicolumn{1}{c|}{Student} &
  \multicolumn{1}{c|}{Teacher} &
  Compression &
  \multicolumn{1}{c|}{ECE} &
  Acc &
  \multicolumn{1}{c|}{ECE} &
  Acc \\ \hline
\multicolumn{1}{|c|}{Base} & \multicolumn{1}{c|}{M0} & \multicolumn{1}{c|}{-}  & 1x   & \multicolumn{1}{c|}{1.81}  & 0.927 & \multicolumn{1}{c|}{0.577} & 0.955 \\ \hline
\multicolumn{1}{|c|}{\multirow{5}{*}{Temp. Scaling}} &
  \multicolumn{1}{c|}{M1} &
  \multicolumn{1}{c|}{M0} &
  8x &
  \multicolumn{1}{c|}{1.762} &
  0.919 &
  \multicolumn{1}{c|}{0.48} &
  0.955 \\ \cline{2-8} 
\multicolumn{1}{|c|}{}     & \multicolumn{1}{c|}{M2} & \multicolumn{1}{c|}{M1} & 16x  & \multicolumn{1}{c|}{1.856} & 0.9   & \multicolumn{1}{c|}{0.44}  & 0.93  \\ \cline{2-8} 
\multicolumn{1}{|c|}{}     & \multicolumn{1}{c|}{M3} & \multicolumn{1}{c|}{M2} & 32x  & \multicolumn{1}{c|}{1.88} & 0.893 & \multicolumn{1}{c|}{0.397} & 0.912 \\ \cline{2-8} 
\multicolumn{1}{|c|}{}     & \multicolumn{1}{c|}{M4} & \multicolumn{1}{c|}{M3} & 64x  & \multicolumn{1}{c|}{1.75}  & 0.845 & \multicolumn{1}{c|}{0.397} & 0.879 \\ \cline{2-8} 
\multicolumn{1}{|c|}{}     & \multicolumn{1}{c|}{M5} & \multicolumn{1}{c|}{M4} & 128x & \multicolumn{1}{c|}{1.897} & 0.861 & \multicolumn{1}{c|}{0.438} & 0.90  \\ \hline
\end{tabular}
}
\end{table}

\begin{table}[!htb]
\centering
\caption{Comparision of IKD+ and IKD using Platt Scaling calibration on EfficientNet-B5}
\label{tab:efficientnetb5}
\resizebox{\columnwidth}{!}{%
\begin{tabular}{cccc|cc|cc|}
\cline{5-8}
                           &                         &                         &      & \multicolumn{2}{c|}{IKD}           & \multicolumn{2}{c|}{IKD+}          \\ \cline{2-8} 
\multicolumn{1}{c|}{} &
  \multicolumn{1}{c|}{Student} &
  \multicolumn{1}{c|}{Teacher} &
  Compression &
  \multicolumn{1}{c|}{ECE} &
  Acc &
  \multicolumn{1}{c|}{ECE} &
  Acc \\ \hline
\multicolumn{1}{|c|}{Base} & \multicolumn{1}{c|}{M0} & \multicolumn{1}{c|}{-}  & 1x   & \multicolumn{1}{c|}{1.89}  & 0.891 & \multicolumn{1}{c|}{0.513} & 0.92  \\ \hline
\multicolumn{1}{|c|}{\multirow{2}{*}{}} &
  \multicolumn{1}{c|}{M1} &
  \multicolumn{1}{c|}{M0} &
  8x &
  \multicolumn{1}{c|}{1.771} &
  0.917 &
  \multicolumn{1}{c|}{0.452} &
  0.915 \\ \cline{2-8} 
\multicolumn{1}{|c|}{}     & \multicolumn{1}{c|}{M2} & \multicolumn{1}{c|}{M1} & 16x  & \multicolumn{1}{c|}{1.625} & 0.918 & \multicolumn{1}{c|}{0.409} & 0.913 \\ \cline{2-8} 
\multicolumn{1}{|c|}{\multirow{3}{*}{Platt Scaling}} &
  \multicolumn{1}{c|}{M3} &
  \multicolumn{1}{c|}{M2} &
  32x &
  \multicolumn{1}{c|}{1.722} &
  0.896 &
  \multicolumn{1}{c|}{0.399} &
  0.896 \\ \cline{2-8} 
\multicolumn{1}{|c|}{}     & \multicolumn{1}{c|}{M4} & \multicolumn{1}{c|}{M3} & 64x  & \multicolumn{1}{c|}{1.781} & 0.822 & \multicolumn{1}{c|}{0.397} & 0.83  \\ \cline{2-8} 
\multicolumn{1}{|c|}{}     & \multicolumn{1}{c|}{M5} & \multicolumn{1}{c|}{M4} & 128x & \multicolumn{1}{c|}{1.78}  & 0.797 & \multicolumn{1}{c|}{0.382} & 0.811 \\ \hline
\end{tabular}%
}
\end{table}
% \usepackage{multirow}
% \usepackage{graphicx}
\begin{table}[!htb]
\centering
\caption{Comparision of IKD+ and IKD using Temperature Scaling calibration on EfficientNet-B5}
\label{tab:efficientnetb5_1}
\resizebox{\columnwidth}{!}{%
\begin{tabular}{cccc|cc|cc|}
\cline{5-8}
                           &                         &                         &      & \multicolumn{2}{c|}{IKD}           & \multicolumn{2}{c|}{IKD+}          \\ \cline{2-8} 
\multicolumn{1}{c|}{} &
  \multicolumn{1}{c|}{Student} &
  \multicolumn{1}{c|}{Teacher} &
  Compression &
  \multicolumn{1}{c|}{ECE} &
  Acc &
  \multicolumn{1}{c|}{ECE} &
  Acc \\ \hline
\multicolumn{1}{|c|}{Base} & \multicolumn{1}{c|}{M0} & \multicolumn{1}{c|}{-}  & 1x   & \multicolumn{1}{c|}{1.918} & 0.91  & \multicolumn{1}{c|}{0.493} & 0.923 \\ \hline
\multicolumn{1}{|c|}{\multirow{2}{*}{}} &
  \multicolumn{1}{c|}{M1} &
  \multicolumn{1}{c|}{M0} &
  8x &
  \multicolumn{1}{c|}{1.922} &
  0.91 &
  \multicolumn{1}{c|}{0.487} &
  0.925 \\ \cline{2-8} 
\multicolumn{1}{|c|}{}     & \multicolumn{1}{c|}{M2} & \multicolumn{1}{c|}{M1} & 16x  & \multicolumn{1}{c|}{1.895} & 0.879 & \multicolumn{1}{c|}{0.491} & 0.886 \\ \cline{2-8} 
\multicolumn{1}{|c|}{\multirow{3}{*}{Temp. Scaling}} &
  \multicolumn{1}{c|}{M3} &
  \multicolumn{1}{c|}{M2} &
  32x &
  \multicolumn{1}{c|}{1.863} &
  0.882 &
  \multicolumn{1}{c|}{0.471} &
  0.845 \\ \cline{2-8} 
\multicolumn{1}{|c|}{}     & \multicolumn{1}{c|}{M4} & \multicolumn{1}{c|}{M3} & 64x  & \multicolumn{1}{c|}{1.821} & 0.853 & \multicolumn{1}{c|}{0.421} & 0.856 \\ \cline{2-8} 
\multicolumn{1}{|c|}{}     & \multicolumn{1}{c|}{M5} & \multicolumn{1}{c|}{M4} & 128x & \multicolumn{1}{c|}{1.79}  & 0.826 & \multicolumn{1}{c|}{0.359} & 0.821 \\ \hline
\end{tabular}%
}
\end{table}
\section{CONCLUSION}
\label{sec:concl}
In this paper, we propose a new model compression algorithm called IKD+ which improves 
precisely with respect to the reliability of predictions obtained by the traditional IKD~\cite{yalburgi2020empirical}. The improvements are introduced due to incorporating temperature- and Platt-scaling in our loss function. The gains are observed in the accuracy of the models obtained. IKD+ also improves the model's reliability by providing more confident predictions than their IKD counterpart.
We apply the IKD+ to address the problem of classifying diseases affecting the retina. The current state-of-the-art in retinopathy is ensemble models consisting of several (in some cases 20) deep convolutional neural networks. These models are resource intensive and impractical for real-world deployment. Our proposal reduces the size of the ensemble models by a factor of 500 while maintaining accuracy comparable to the state-of-the-art. Our proposal is also more reliable, offering a better trade-off between accuracy, ECE and size than shifting between different versions of EfficientNets.  
\begin{figure}[!htb]
    \centering
    \includegraphics[width=0.8\linewidth]{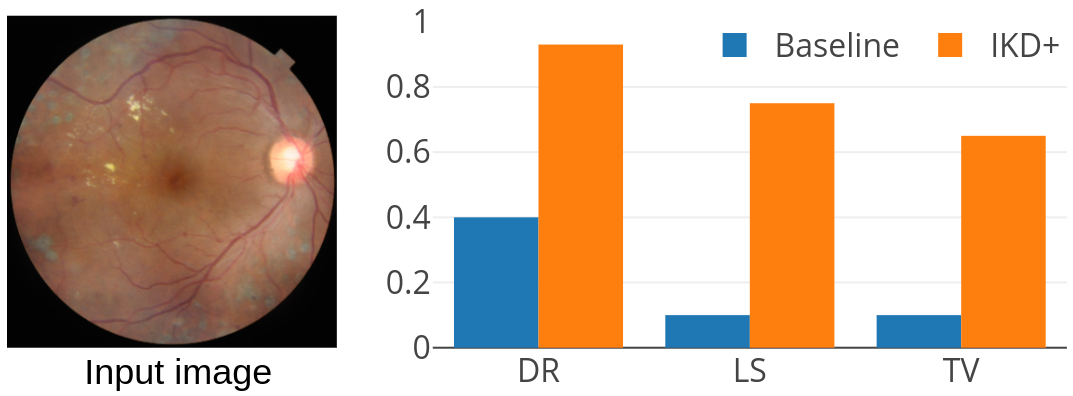}
    \caption{ Bar plots indicating the probability of classifying an image as diabetic retinopathy (DR), laser scar (LS) and tortuous (TV) using baseline and IKD+ model.}
    \label{fig:preds}
\end{figure}

%\vfill
%\pagebreak
% \bibliographystyle{IEEEbib}
% %\bibliography{strings,refs}
% \bibliography{references}

\end{document}